\documentclass[a4paper,conference]{IEEEtran}

\usepackage{graphicx}
\usepackage{caption}
\usepackage{blindtext}
\usepackage{xcolor}
\usepackage{multicol}
\usepackage{lipsum}
\usepackage{float}
\usepackage{svg}
\usepackage{pdfpages}
\usepackage{mathtools}

\usepackage{esvect}
\usepackage{dblfloatfix}
\usepackage{subfig}
\usepackage[export]{adjustbox}
\usepackage[noadjust]{cite}

\restylefloat{table}

\begin{document}
\title{Political Posters Identification with Appearance-Text Fusion}

\author{
    \IEEEauthorblockN{Xuan Qin\IEEEauthorrefmark{1}, Meizhu Liu\IEEEauthorrefmark{2}, Yifan Hu\IEEEauthorrefmark{2}, Christina Moo\IEEEauthorrefmark{2}, Christian M. Riblet\IEEEauthorrefmark{2}, \\Changwei Hu\IEEEauthorrefmark{2}, Kevin Yen\IEEEauthorrefmark{2}, Haibin Ling\IEEEauthorrefmark{1}}
    \IEEEauthorblockA{\IEEEauthorrefmark{1}Stony Brook University
    \\\{xuqin, hling\}@cs.stonybrook.edu}
    \IEEEauthorblockA{\IEEEauthorrefmark{2}Yahoo! Research
    \\\{meizhu, yifanhu, cmoo, riblet, changweih, kevinyen\}@verizonmedia.com}
}

\maketitle

\begin{abstract}
In this paper, we propose a method that efficiently utilizes appearance features and text vectors to accurately classify political posters from other similar political images. The majority of this work focuses on political posters that are designed to serve as a promotion of a certain political event, and the automated identification of which can lead to the generation of detailed statistics and meets the judgment needs in a variety of areas. Starting with a comprehensive keyword list for politicians and political events, we curate for the first time an effective and practical political poster dataset containing 13K human-labeled political images, including 3K political posters that explicitly support a movement or a campaign. Second, we make a thorough case study for this dataset and analyze common patterns and outliers of political posters. Finally, we propose a model that combines the power of both appearance and text information to classify political posters with significantly high accuracy.
\end{abstract}

\IEEEpeerreviewmaketitle

\section{Introduction}

\begin{figure}
\centering
\includegraphics[width=0.12\textwidth,keepaspectratio, frame]{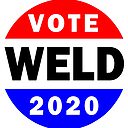}
\includegraphics[width=0.12\textwidth,keepaspectratio, frame]{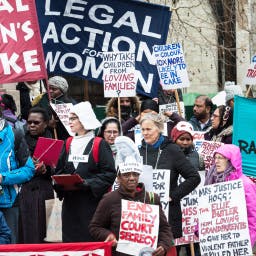}
\includegraphics[width=0.12\textwidth,keepaspectratio, frame]{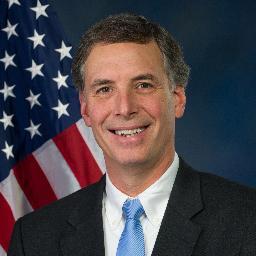} \\
\includegraphics[width=0.12\textwidth,keepaspectratio, frame]{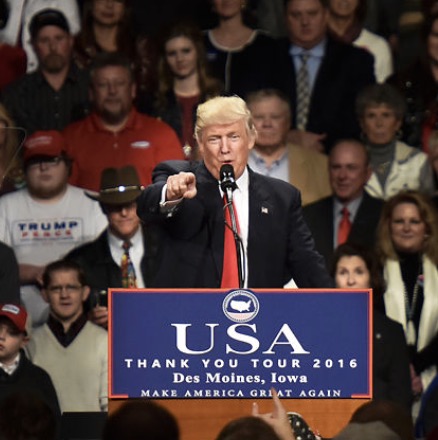}
\includegraphics[width=0.12\textwidth,keepaspectratio, frame]{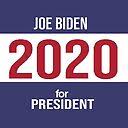}
\includegraphics[width=0.12\textwidth,keepaspectratio, frame]{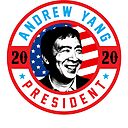} \\
\includegraphics[width=0.12\textwidth,keepaspectratio, frame]{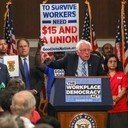}
\includegraphics[width=0.12\textwidth,keepaspectratio, frame]{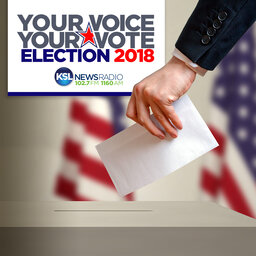}
\includegraphics[width=0.12\textwidth,keepaspectratio, frame]{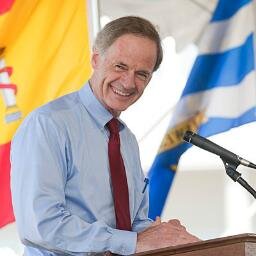}
\caption{Various political images containing both appearance features and text information.}
\label{poticalimage}
\end{figure}

Online political advertisements have been a controversial and sensitive issue. In certain states, the companies, by law and regulation, need to disclose the information of the advertisements that originated from different political campaigns. Aside from such disclosure, some companies straight out ban political ads as part of their policy. In order to follow the regulations and policies, precise identification of a political poster is required. In the agreement, an advertising platform may wish to enforce the client to mark their own entries as political, but a precise systematic validation is still required and is not trivial. Moreover, advertisements being distributed to different platforms and publishers may lose such critical context in the process of transmission, while other ads simply straight out do not contain such information. Since the volume is infeasible for a thorough manual review of every entry, automatic classification of political posters becomes a helpful aid to the policy team. The image of the posters may be a scene or a designed image for certain political candidates or movements, alongside slogans and signs. Therefore the classification task can be designed to consume these features. Fig. \ref{poticalimage} shows various political images containing both appearance features\footnote{in this paper, we refer to image features that are not explicitly extracted into texts as \em{appearance features}} and text information.

These political posters usually come with a variety of objects to focus on. For instance, a congressman portrait would give object labels such as flags and suits. Text labels such as “2020” and “Vote” can be found in an election poster, which directly indicates the purpose within the image. Object appearance features and textual information are usually considered as different areas and studied differently, however in our task of political image classification, we are aiming to create a model that simultaneously uses them for the classification task.

\begin{figure*}[!h]
\centering
\includegraphics[width=0.9\textwidth,keepaspectratio]{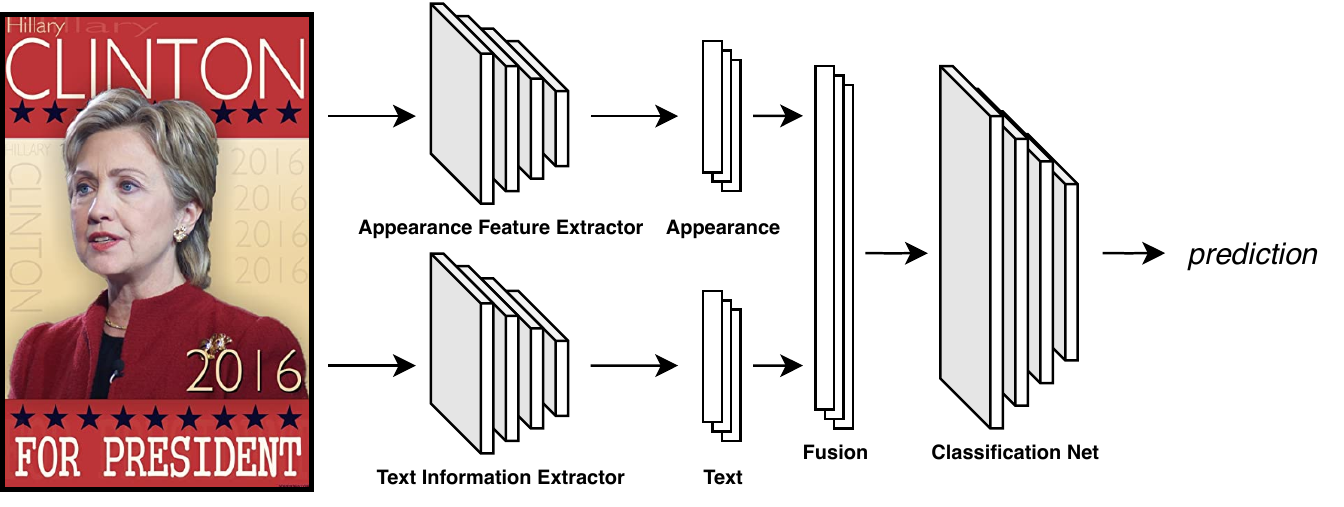}
\caption{Our method: fusion of appearances and texts.}
\label{intro_diag}
\end{figure*}

In this paper, our key contributions are:

\begin{itemize}
    \item We provide the first dataset of political poster images for further research.
    \item We propose the very first approach to identify political poster images with high accuracy by combining both appearance features and text information
\end{itemize}

To curate the dataset, we first create a list of keywords from different political areas with a set of practical suffixes to accurately search for political posters. The dataset is built with a weighted number of images per category with respect to the possibilities of political posters. 

To classify political posters effectively, we propose a model that fuses both appearance and textual features.  
As shown in Fig. \ref{intro_diag}, considering the various appearance and textual features of political posters, two separate models are made use of according to each specific task: An appearance feature detector model to find pixel patterns and a text recognition model to produce text vectors. Having both appearance patterns and text information, we concatenate them in a weighted distribution to generate a single long vector and build a 3-layer fully-connected neural network to make predictions from the vector.

\section{Related Work}

The search of political-related research leads us to unexpected results from the past decades. A tremendous amount of work has been devoted into the area of political advertising, discussing the power and influence of it \cite{political_poster_ref_1, political_poster_ref_2, political_poster_ref_3, political_poster_ref_4, political_poster_ref_5, political_poster_ref_6, political_poster_ref_7, political_poster_ref_8, political_poster_ref_9}. Rosenberg et al. \cite{1} show us how detailed image features can influence the election. With the power of deep learning, some related research has covered the hidden information that lies inside political images. The deep neural network is used in \cite{2} to analyze the frequency of identified objects found in political images, as well as black representations among white house members’ uploaded images on social media, using a dataset of photos from Facebook pages of members of the U.S. House and Senate. From their results, we peek the power of deep neural networks that can be brought to the field of political image recognition and classification. Some research \cite{3} draws attention to how political symbols can play as an important role in political communications. Wiedemann and Gregorcite \cite{4} propose a workflow of active learning that shifts between human operations and automatic machine learning to solve the uneven distribution of characteristic language structures in the field of political manifestos.

There is a large range of discussions and research on political images and advertisements, while all of them address the importance of the power and influence behind political images, few of them mention the identification with deep neural networks. And in the field of computer vision, rare research is done combining the recognition of texts and appearances of objects. A practical model \cite{fusion1} introduces the method of fusing images and texts to classify pictures by acquiring texts from the webpages where images are from, but it can not respond to scenarios where there are only images as input. In \cite{fusion2}, the authors select a different strategy by recognizing texts directly from images, and for image features, they use the bag-of-words method and derive probabilities from it. However, an SVM-based classifier is in the final stage as the classification method, which is proven to perform worse than deep neural networks in contemporary problems with high-dimensional complexity.

\section{Our Method}

\subsection{Dataset}

We create the first political image dataset by searching diversified political keywords and manually labeling the searching results. This dataset is built with 2k keywords, including names of popular politicians (e.g., U.S. senators) and titles of party ideologies and movements. We create the keywords by acquiring the lists of names of politicians from these fields and adding several general keywords that practically bring with political posters results. Since the search result of a specific keyword is sorted by relevance and bottom images are likely to be irrelevant and random, we also limit the number of images per keyword to be relatively low corresponding to the field a keyword belongs to as well as its suffix.

We analyze the preliminary dataset made from keywords with different suffixes in different categories. However, it comes to our conclusion that, despite how practical the keywords fed into search engines are, the result images are not certainly political posters with possible promotion intentions. And corresponding to the field of keywords, e.g. U.S senators, the proportion of political images varies in the results. We therefore slightly adjust the suffixes after several tryouts and weight each type of keywords and suffix such that there would be more posters images that are actually advertisements in this dataset. There are 4 types of suffixes we add to the original keywords, and as the result of annotations shows, these suffixes are proven to be practical and effective in terms of increasing the percentage of target political posters in the search results. 

\begin{table}[!thb]
	\centering
	\caption{Number of keywords in each category.}
	\begin{tabular}{lc}
		\hline
		Category & \# of keywords \\
		\hline
		Popular Politicians & 39 \\
		International Politicians & 338 \\
		Candidates & 1301 \\
		U.S. Senators & 100 \\
		U.S. House & 433 \\
		Parties Ideologies & 170 \\
		Paid for by & 34 \\
		California Propositions & 17 \\
		General & 34 \\
		Cartoons & 1 \\
		\hline
	\end{tabular}
\end{table}

\begin{table}[!tbh]
	\centering
	\caption{Number of images per keyword in Party Ideologies.}
	\begin{tabular}{lc}
		\hline
		Suffix & \# of images per keyword\\
		\hline
		w/ ad & 5 \\
		w/ poster & 20 \\
		w/ election poster & 40 \\
		w/ political poster & 40 \\
		\hline
	\end{tabular}
\end{table}

The final dataset satisfies the requirement of our goal well. After labeling, it has 13k images that are mostly political-related and 3k of them are identified as political posters with the explicit purpose of promoting a campaign or movement. The proportion of it also meets the real scenario where most political images are simply neutrally demonstrating a political event.

To simulate possible real-life scenarios for the test session, we also add 8.2k natural images from ImageNet \cite{imagenet_cvpr09} and 1.8k non-political poster images, which are crawled using keywords such as "concert posters". We believe the diversity of our dataset can provide abundant possibilities for testing and validation purpose and it will also give a concrete proof that our method works well in real cases.

\subsection{Overview of Structure}

Our method benefits from modern deep neural networks. We acquire a appearance vector and a text vector from corresponding extractors:
\begin{equation}
v_{a} = G(\mathbf{I})
\end{equation}
\begin{equation}
v_{t} = F_{encoding}(H(\mathbf{I}))
\end{equation}
where $v_{a}$ is the appearance vector extracted by the appearance feature extractor, a convolutional neural network, $G$ with $\mathbf{I}$ as the input RGB image. $v_{t}$ is the text vector and $H$ is the text information extractor. The $F_{encoding}$ we use here is one-hot encoding which ensures a consistent dimension despite of the varying number of words in each image. We then concatenate these two vectors to create a longer vector:
\begin{equation}
v_{long} = v_{a} + k \times v_{t}
\end{equation}
where $k$ is a scalar to multiply the text vector, and then feed this long vector to the classifier, which is a fully-connected network followed by a Sigmoid function that remaps the output so that we can make it range between 0 and 1. We set a decision boundary at 0.5 to produce binary predictions. Binary cross-entropy is used when calculating the loss.

\subsection{Appearance Features and Texts}

\begin{figure*}[t]
\centering
\includegraphics[width=\textwidth,keepaspectratio]{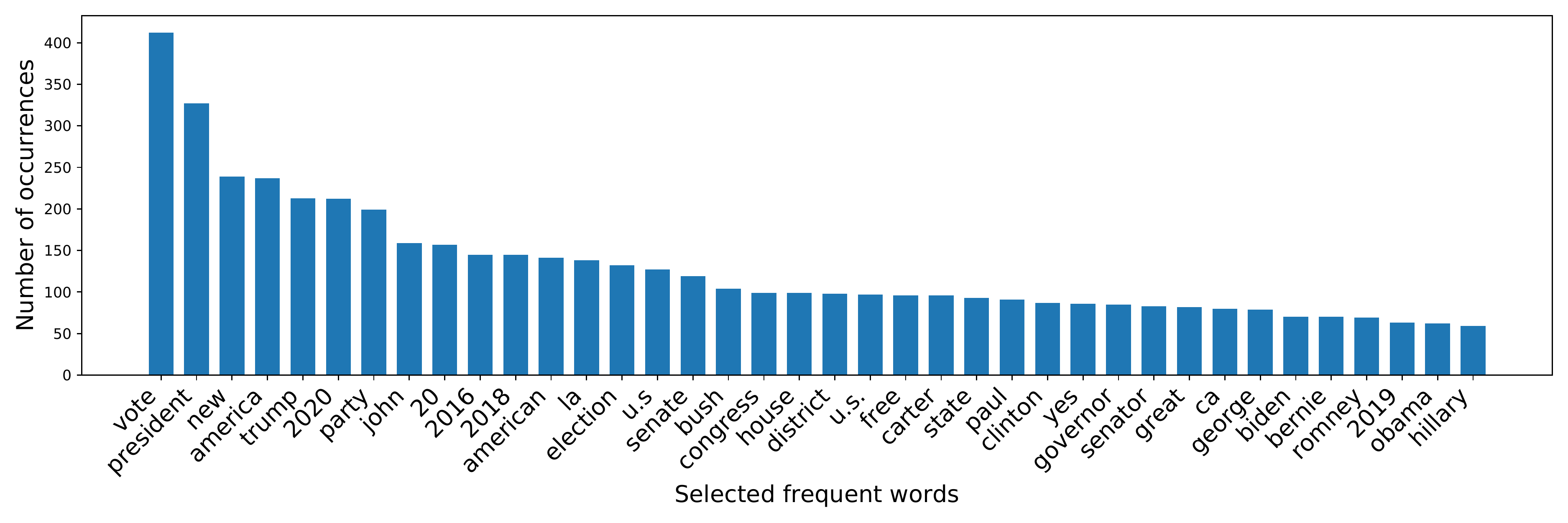}
\caption{The histogram we build upon the result of the text annotation result. The length of the text table that we store in the annotation stage is 500k, which means at most 500k individual words are saved in the list. After sorting the number of how many times a word appears in the dataset, we take the top 50k of them. This figure shows a set of selected political words. Irrelevant words, such as article words, because of their small amount compared to the length of vector we use in experiments, are not removed from the list.}
\end{figure*}

Deep neural networks construct a complicated model to learn high-level patterns of images, e.g. the pattern of flags and suits. We take this advantage to extract appearance features that lie inside political poster images. We fine-tune all pre-trained models available in PyTorch with our dataset by modifying the last fully-connected layer to produce a single output and then select the model with the highest accuracy: ResNeXt \cite{resnext}, specifically, \textit{resnext50\_32x4d}.

ResNeXt proposes the combination of ResNet and split-transform-merge method from Inception. The \textit{resnext50\_32x4d} model from PyTorch has 50 layers with cardinality as 32 and bottleneck width as 4-$d$. We delete the last layer of the pre-trained model from PyTorch, such that this feature extractor produces an appearance vector of 2048-$d$, which, as our result suggests, is efficient to present the necessary high-level image patterns to classify political posters.

Appearance features alone are not sufficient for this task, as political posters almost always come with texts, and they can significantly help with our judgment of whether they are promotional or not.

Texts serve as a crucial role in representing the ideas behind a certain campaign or sometimes become the slogan of a movement. Extraction of these texts is necessary if we want to identify an image as a political poster at the first step, then it comes to the understanding of words, and determining whether it is related to political advertising or not. We believe images are very likely to be political if certain words appear in them.

To validate this idea, we build a word histogram, shown in Fig. 3, using our dataset. Benefiting from highly developed text recognition methods, we can directly derive texts from scenes. An industry-level annotator, Google Cloud Vision API, is used here to ensure the precision such that the result can be considered as ground truth. Although the histogram is heavily interfered with by many other random texts that appear inside images, it is clear that a number of political-sensitive words make their way to the top of the histogram, as expected.

Instead of making use of other features like heatmaps generated from text recognition models, we directly build a vector list of frequent words that can appear in political posters based on the histogram. For each element in the list, it stands for a particular word, e.g. 'vote', and the value of each element in the list is exactly how many times that word appears in the image. Because of the unknown number of words that can appear in an image, we have to use one-hot encoding for the text vector to assure dimension consistency. Despite the long length of the vector, most of its values are 0's, which makes it easy to encode and store them in a reasonably small size.

\subsection{Fusion}

In order to combine the appearance vector and the text vector, it is necessary to determine the weight of a text vector. Compared to appearance features, we expect the role of text vectors not to be significantly more dominant. As given in a particular scene, the images may give out a set of highly frequent texts, for instance, ‘vote’, but does not provide any advertisement for any campaign or individual. This situation appears in a great many political cartoon and figure images. To solve this issue, we set a weight factor for the text vector. The text vector is scaled before the fusion with the appearance vector, such that there is a well-chosen trade-off between image appearances and texts.

In terms of the fusion network, we find adding more dense layers would slightly contribute to higher overall accuracy in the test, meaning that the true complexity of this classification task can be relatively higher than expected. As we have prepare enough inputs for the fusion model, we believe there is no need to add more fancy layers to the network itself. We simply build a network with 1 dense layer, or 3 dense layers with 1 ReLu layer after the first each two. Despite having more layers, e.g., 5 layers, will enhance the result a little bit, it is more acceptable to lower down the number of layers for less computational costs.

The output from the neural network is a single value that will be fed to a Sigmoid function. The Sigmoid function simply remaps the output to the range between 0 and 1, which has to be done before the binary cross-entropy criterion, as we set our target value of political posters as 1 and the others as 0. With Adam optimizer, the training process finishes rapidly, benefiting from the simplicity of structure.

\section{Experiments}

We create a dataset of 13k labeled images crawled with keywords that are all politics-related, and another 10k of natural images and other poster images, e.g., concert posters. The first 13k images are classified into political poster images, political but not advertisement images and off-topic images, where the number of off-topic images is only 1.5k, and all the rest are identified as political images. We believe this dataset generally can provide the same scenarios as real-life situations where most political images, among other types of photos, are about neutral statements or interview photographs and have no bias in favor of any side of a competition.

With the dataset, first, we test the result on the first 13k dataset using ResNeXt alone, which reaches 85.1\% accuracy. As we analyze the false predictions of this test, it appears that ResNeXt can successfully extract the most common appearance features of political images. The false positives are most likely to be political images with congressmen or flags of a nation, etc. This validates our initial idea that political poster image recognition requires the combination of texts and objects appearances.

\begin{figure}[h!]
\centering
\subfloat[False positive.]{%
\includegraphics[height=0.12\textheight, frame]{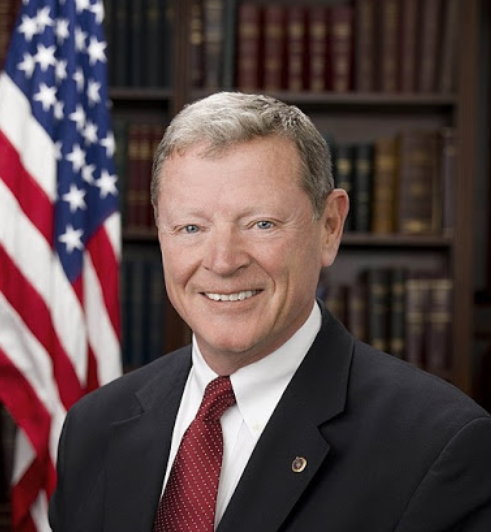}}
\hspace*{8px}
\subfloat[False negative.]{%
\includegraphics[height=0.12\textheight, frame]{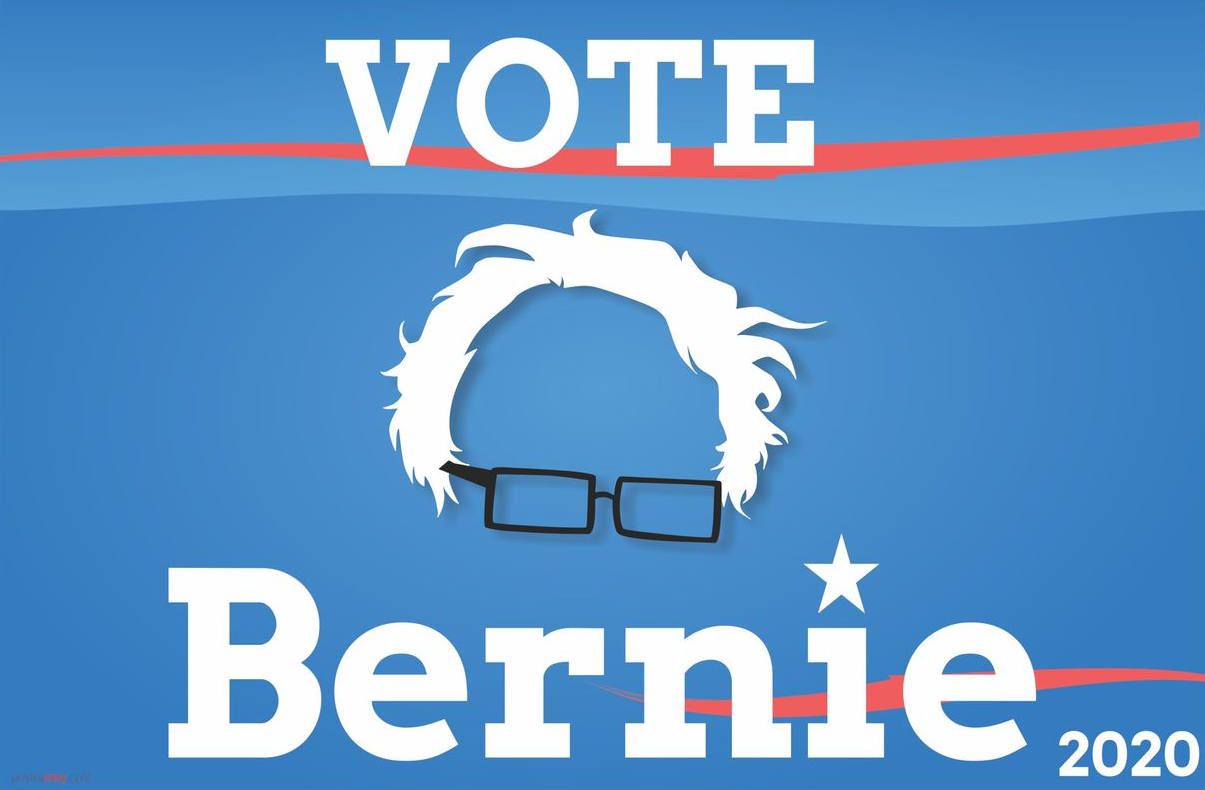}}
\caption{In (a), the suit and U.S. flag are correctly recognized by the model, which however, should not lead to the conclusion that it is a political poster image. In (b), appearance features are no longer important since the texts play a much more crucial role in representing the poster.}
\end{figure}

We then remove the last layer of \textit{resnext50\_32x4d} and feed it the entire dataset to generate appearance features of each image. The feature vector of each image is $2048\times1$. Considering the objects that appear in political images are in a rather small range according to our analysis of the dataset, we believe the 2048-dimensional vector is sufficient in this task.

For text vectors, we build a vacant dictionary that is 500k in length, accommodable for general memory size limit. Then, we use the Google Cloud Vision API to annotate the dataset. Each word in each image will fill in the dictionary, and its value is the number of its occurrences in that image. The 500k length is handy for our 23k dataset, since the limit length of dictionary is never reached. We sort the dictionary by value and leave only the top 50k. The annotations are considered as ground truth in our experiment. To ensure this assumption, 300 randomly picked images are human-labeled and are compared to the result from Google Cloud Vision API, and there is almost no difference at all. In terms of industrial implementation of our method without using an online service, we believe using any state-of-the-art text recognition model can also generate a same close result here, as a benefit of the fact that most words appear on posters are rather large and clear to identify. Additionally, because words from non-poster images are more difficult to recognize, it sometimes leads to the failure of recognition and their values in text vectors remain as 0's, so that they are less likely to be classified as political posters.

We also conduct an experiment with different dimensional text vectors to see if we can save memory and training time. We build a histogram using the result of text annotation and sort by histogram then take the both top 3k and top 50k most frequent words as the text vectors to test separately. The results in the test sessions lead to our conclusion that the length of the text vector before fusion is depending on which test set we have. In different scenarios, the length can be longer or shorter accordingly to produce the best accuracy rather than over-fitting or under-fitting.

\begin{figure}
\centering
\includegraphics[width=.4\textwidth]{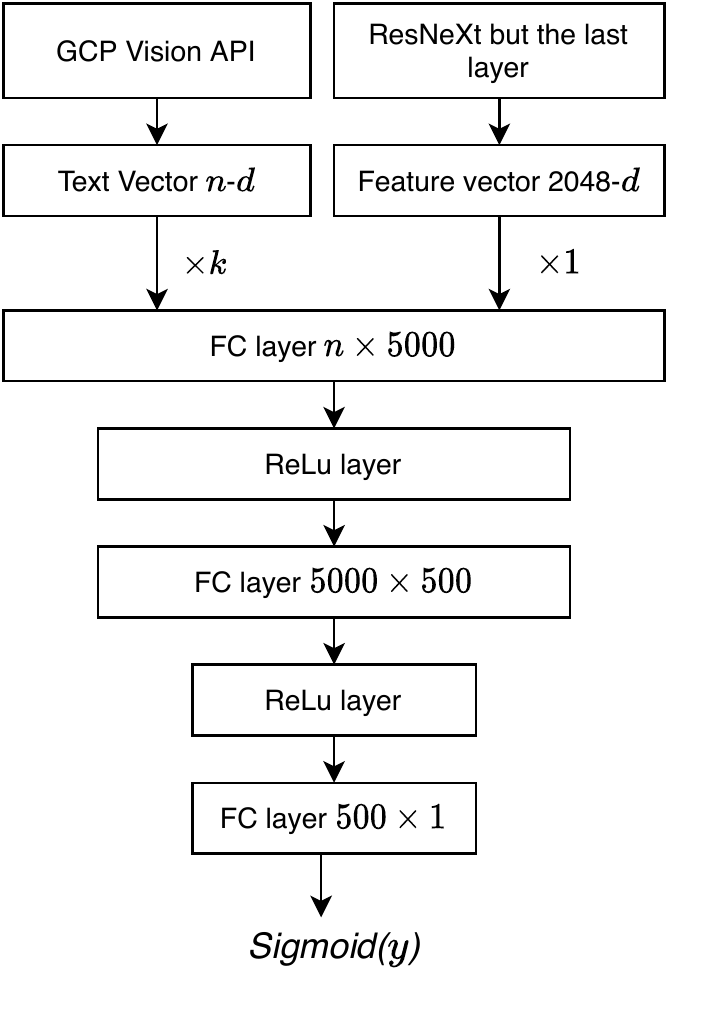}
\caption{Diagram of the network. $n$ stands for the length of text vector, we test both $n=3000$ and $n=50000$. To use this offline, one can easily replace Google Cloud Vision API with any state-of-the-art text recognition model.}
\label{diag}
\end{figure}

Our best classification model is of three fully-connected layers, with ReLu activation functions after the first two. We test a set of different settings, e.g. different layers setups and find this model not necessarily the most time-efficient but the most accurate in many cases while saving memory. As shown in Fig. \ref{diag}, to combine the appearance vector and the text vector, we simply concatenate them to a single higher dimensional vector. The appearance vector is $2048\times1$ and the text vector is $n\times1$. After concatenation the input vector of each image is $(n+2048)\times1$. The $k$ stands for the factor that the text vector will multiply before the concatenation, which is 0.5 in our tests.

\begin{table*}[ht]
\centering
\caption{Fusion vs no fusion where $n=3000$. D stands for dummy classifier, R stands for ResNeXt, I stands for Inception, T stands for text vector, RT and IT stands for the fusion of appearance and text vector, and 3-L stands for a 3-layer classifier. The last layer of \textit{resnext50\_32x4d} is modified to produce a single output. Same applies to \textit{inception\_v3} and we only use the main output from it. Unless otherwise specified, all models use 1-layer classifiers. All the first 4 datasets are of 13k images in total. The 5th dataset setup is meant to validate the performance with a balanced negative/positive ratio, however, in real-life scenarios, this ratio would be much higher, more like that of the first 4 setups.}
\begin{tabular}[t]{||c | c c c c c | c |c c c c c c c ||}
\hline
\multicolumn{6}{||c|}{Dataset Setup} & \multicolumn{8}{c||}{Model Setup ($n=3000$)}\\
\hline
Index & Political & Positive & Off-Topic & Natural & Poster & D & R & I & T & RT & IT & RT 3-L & IT 3-L \\
\hline\hline
1 & 8.5k& 3k& 1.5k& - & -    & 77.6 & 85.1 & 84.6 & 83.9 & \textbf{88.2} & 86.3 & 87.2 & 86.3 \\
2 & 5.5k& 3k& 1.5k& 3k& -    & 77.6 & 90.7 & 89.8 & 89.3 & 92.1 & 91.1 & \textbf{92.4} & 91.6 \\
3 & 3.5k& 3k& 1.5k& 5k& -    & 77.5 & 92.4 & 91.6 & 90.7 & 93.5 & 93.0 & \textbf{93.7} & 92.9 \\
4 & -& 3k& -& 8.2k& 1.8k    & 75.6 & 93.9 & 92.5 & 94.4 & 96.7 & 95.5 & \textbf{96.8} & 96.1 \\
\hline
5 & -&    3k& 1.5k& -&  1.8k & 53.0 & 81.5 & 78.5 & 86.8 & 88.2 & 87.1 & \textbf{90.1} & 87.9 \\
\hline
\end{tabular}
\end{table*}

For a better validation of our method, we test 7 models in 5 different data setups. 7 models are ResNeXt alone, Inception alone, text vector alone, ResNeXt and text vector fusion with 1-layer classifier, Inception and text vector fusion with 1-layer classifier, plus two models that each has a 3-layer classifier. Inception models are for comparison against ResNeXt as the feature extractor. For text vectors, we mainly test $n=3000$, since it's more practical in terms of memory size, and $n=50000$ is also tested in some experiments.

The complete dataset we have consists of 8.5k political images, 3k political poster images as positive samples, 1.5k off-topic images from the crawling of political posters, 8.2k natural images from ImageNet and 1.8k normal poster images. It is essential how we build up different subsets of this dataset for training and testing to re-implement real-life scenarios that requires political poster identifications from a group of different unknown type images. We decide to build several dataset setups to validate our methods in a comprehensive manner. Some of them are heavily appearance feature based, for instance, where we add more natural images in a certain setup, while other setups may have more text-based political images or poster images. We compare the performance of all our models in all these dataset setups and eliminate possible randomness by running K-Fold 5 and averaging results.

\begin{table}[t!]
\centering
\caption{Tests for $n=3000$ and $n=50000$, where $n$ stands for the length of text vectors. Model is fusion of ResNeXt and text vectors, with a 3-layer classifier.}
\begin{tabular}[t]{||c | c c ||}
\hline
Dataset Index& $n=3000$&$n=50000$\\
\hline\hline
3&93.7&94.4\\
4&96.8&97.2\\
\hline
5&90.1&90.3\\
\hline
\end{tabular}
\end{table}

The results suggest the fusion model compensates deficiencies of appearance model and text vector model by combining their respective advantages. In the most difficult dataset setup, where there are 8.5k political images to intervene identifications of political posters, our method still has a reasonable accuracy. With the decreasement of political images and the increasement of other types of images, our approach reaches even higher accuracy with balanced precision and recall rates. The learning of our models is very stable in all scenarios, as it gradually converges from early to late epochs. The convergence speed varies between different setups, but in general, our fusion models converge much faster than appearance-only and text-only models, which in average, require 2 times more epochs than fusion models to converge. Higher number of layers of classifiers also helps with the speed.

We believe our method is proven to be reliable in terms of generalization without over-fitting, as given this result, 5 tests in K-Fold 5 have very little variance in distribution with each model stably converging to a low loss that is close to each other. The $n$-dimensional text vector is proven to be practical as the results suggest. The combination of text vector and appearance vectors brings the final improvement to this task as it successfully identifies the non-political-poster images without or without texts.

\section{Conclusion}

This paper establishes for the first time a political poster dataset and contributes the first effective approach to classify political images. This method consists of an appearance feature detector, a text classifier and a fusion model, which successfully combines the utilizations of both images and texts in political posters. Our fusion approach reaches a higher accuracy with even faster convergence speed. It is stable in training and provides excellent predictions while requiring minimal amount of configuration and a small computational cost because of its structural simplicity.

Further research can be conducted with the training setup. In our experiment, the uncertainty of the text vector length has been an issue as we do not know what length would be the most efficient and accurate. While $n=3000$ performs well in most cases, we have not investigated other $n$ values extensively. And because of the later concatenation and neural network processing, the text vector must be stored as float type, and if the ideal vector is very long, a huge float vector for a single image is not considered memory friendly. This also leads to an increase of the complexity of the neural network model, which consumes more time to back-propagate. 
Further research can be done on the trade-off between accuracy and computational efficiency.

\bibliographystyle{IEEEtran}
\bibliography{polimg_conference.bib}

\end{document}